\documentclass[conference]{IEEEtran}
\IEEEoverridecommandlockouts
\usepackage{cite}
\usepackage{amsmath,amssymb,amsfonts}
\usepackage{algorithmic}
\usepackage[dvips]{graphicx}
\usepackage{textcomp}
\usepackage{xcolor}

\usepackage{subfigure}
\usepackage{url}
\usepackage{listings}
\usepackage{algorithm}
\usepackage{algorithmic}
\usepackage{multirow}

\newcommand{\bvec}[1]{\mbox{\boldmath $#1$}}

\def\BibTeX{{\rm B\kern-.05em{\sc i\kern-.025em b}\kern-.08em
    T\kern-.1667em\lower.7ex\hbox{E}\kern-.125emX}}
\begin{document}

\title{An Object Detection by using Adaptive Structural Learning of Deep Belief Network
\thanks{\copyright 2019 IEEE. Personal use of this material is permitted. Permission from IEEE must be obtained for all other uses, in any current or future media, including reprinting/republishing this material for advertising or promotional purposes, creating new collective works, for resale or redistribution to servers or lists, or reuse of any copyrighted component of this work in other works.}
}

\author{\IEEEauthorblockN{Shin Kamada}
\IEEEauthorblockA{\textit{Graduate School of Information Sciences} \\
\textit{Hiroshima City University}\\
3-4-1, Ozuka-Higashi, Asa-Minami-ku,\\
Hiroshima, 731-3194, Japan\\
Email: da65002@e.hiroshima-cu.ac.jp}
\and
\IEEEauthorblockN{Takumi Ichimura}
\IEEEauthorblockA{\textit{Faculty of Management and Information Systems} \\
\textit{Prefectural University of Hiroshima}\\
1-1-71, Ujina-Higashi, Minami-ku,\\
Hiroshima, 734-8559, Japan\\
Email: ichimura@pu-hiroshima.ac.jp}
}

\maketitle

\begin{abstract}
  Deep learning forms a hierarchical network structure for representation of multiple input features. The adaptive structural learning method of Deep Belief Network (DBN) can realize a high classification capability while searching the optimal network structure during the training. The method can find the optimal number of hidden neurons for given input data in a Restricted Boltzmann Machine (RBM) by neuron generation-annihilation algorithm. Moreover, it can generate a new hidden layer in DBN by the layer generation algorithm to actualize a deep data representation. The proposed method showed higher classification accuracy for image benchmark data sets than several deep learning methods including well-known CNN methods. In this paper, a new object detection method for the DBN architecture is proposed for localization and category of objects. The method is a task for finding semantic objects in images as Bounding Box (B-Box). To investigate the effectiveness of the proposed method, the adaptive structural learning of DBN and the object detection were evaluated on the Chest X-ray image benchmark data set (CXR8), which is one of the most commonly accessible radio-logical examination for many lung diseases. The proposed method showed higher performance for both classification (more than 94.5\% classification for test data) and localization (more than 90.4\% detection for test data) than the other CNN methods.
\end{abstract}


\section{Introduction}
Recently, deep learning methods are extensively applied to various fields of science and engineering for the training large volume data \cite{Bengio09, Quoc12, webmarket2016}. Some current prominent techniques have been extensively and rapidly studied for the various kinds of problems such as classification, object detection, time-series prediction, and so on \cite{Mohammadi18}. Their new architectures of CNN (Convolutional Neural Network) such as AlexNet \cite{AlexNet}, GoogLeNet \cite{GoogLeNet}, VGG16 \cite{VGG16}, and ResNet \cite{ResNet} have been developed \cite{Russakovsky15}.

In deep learning, trial and error to find the optimal network structure for given input data will be required and the development of new structure is a difficult task even for experienced designers. Due to the computational cost or resource for big data representation and its analysis of deep learning, the investigation of possibility for all parameter sets is impracticable. The transfer learning is famous method of re-use of the trained network with high classification capability such as GoogLeNet, VGG16, or ResNet for a new problem, because it is easier to construct a network structure for a new data. However, in order to achieve powerful classification capability, we consider that the representation of new data requires the specified network structure with the covered feature of data space. The learning method by using the pre-trained model cannot express the characteristic data representation. We have proposed the adaptive structural learning method of DBN \cite{Kamada18_Springer}. The method has an outstanding function of determination for the structure of RBM \cite{Hinton06,Hinton12} which has the self-organized algorithm by hidden neuron generation and deletion during learning phase. The number of RBM layers is also automatically defined by the layer generation method \cite{Kamada16_SMC, Kamada16_ICONIP, Kamada16_TENCON}. The adaptive structural learning of DBN method shows the highest classification capability for image recognition of the benchmark data sets such as MNIST \cite{LeCun98a}, CIFAR-10, and CIFAR-100 \cite{CIFAR10}. The classification accuracy for training data sets was almost 100\% and 99.5\%, 97.4\%, and 81.2\% for test cases, respectively \cite{Kamada18_Springer}.

In this paper, a new object detection method for the DBN architecture is proposed for localization and category of objects. The method is a task for finding semantic objects in images as Bounding Box (B-Box). The basic idea is similar to famous object detection methods of CNN such as R-CNN \cite{RCNN}, YOLO \cite{YOLO}, and SSD \cite{SSD}, but the method of CNN cannot be applied to DBN architecture because of the image feature of convolutional filter. Moreover, the method of CNN also estimates the probability of semantic objects in an image as a continuous heatmap. On the contrary, the proposed method can represent discrete heatmap, since the hidden neuron takes binary value \{0, 1\}.

The proposed method was evaluated on the Chest X-ray image benchmark data set (CXR8) \cite{CXR8}, which is one of the most commonly accessible radio-logical examination for many lung diseases. Compared with the result of transfer learning with well-known CNN methods, the proposed method showed higher performance for both classification (more than 94.5\% classification for test data) and localization (more than 90.4\% detection for test data) than the other CNN methods.

The remainder of this paper is organized as follows. In section \ref{sec:adaptive_dbn}, basic idea of the adaptive structural learning of DBN is briefly explained. Section \ref{sec:dbn_detection} gives the description of the proposed object detection algorithm and generation method of heatmap by using the trained DBN network. In section \ref{sec:exe}, the effectiveness of our proposed method is verified on CXR8. In section \ref{sec:conclusion}, we give some discussions to conclude this paper.

\section{Adaptive Learning Method of Deep Belief Network}
\label{sec:adaptive_dbn}
This section explains the traditional RBM \cite{Hinton12} and DBN \cite{Hinton06} to describe the basic behavior of our proposed adaptive learning method of DBN.

\subsection{Restricted Boltzmann Machine}
\begin{figure}[bt]
\centering
\includegraphics[scale=0.8]{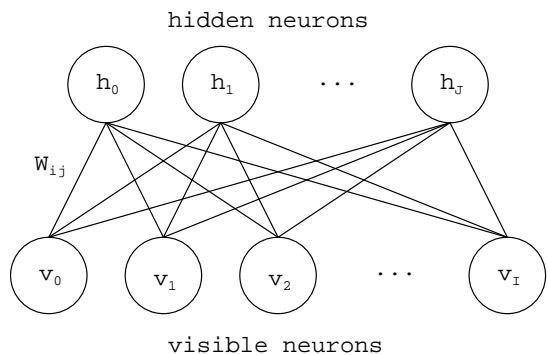}
%\vspace{-3mm}
\caption{Network structure of RBM}
\label{fig:rbm}
\end{figure}

A RBM \cite{Hinton12} is a stochastic unsupervised learning model. As shown in Fig.~\ref{fig:rbm}, RBM has the network structure with two kinds of layers $\bvec{v} \in \{0, 1 \}^{I}$ and  $\bvec{h} \in \{0, 1 \}^{J}$, and three parameters $\bvec{\theta}=\{\bvec{b}, \bvec{c}, \bvec{W} \}$. There are two important properties in RBM, one is the neuron is represented by binary value. The other is there is no connection among same layers. These properties enable each hidden neuron to learn independent feature of given input patterns.

Since RBM is a stochastic model, the optimal parameter $\bvec{\theta}$ for given input can be found by a maximum likelihood estimation method. Contrastive Divergence (CD-$k$) \cite{Hinton02} uses two tricks to speed up the Gibbs sampling method as the most popular RBM learning method.

\subsection{Deep Belief Network}
\label{subsec:dbn}
DBN \cite{Hinton06} is multiple layers of graphical model for stochastic unsupervised learning. The most popular DBN is a form of layers of two or more RBMs for the pre-training phase. The output patterns of hidden neurons at the $l$-th RBM can be seen as the next input at the $(l+1)$-th RBM. Successively, the trained DBN calculates a feed-forward network for the fine-tuning. In the supervised learning for classification task, the final output layer works to calculate the output probability $y_k$ for a category $k$ by Softmax method.

%Deep Belief Network (DBN) \cite{Hinton06} is a stacking box for stochastic unsupervised learning by hierarchically building several pre-trained RBMs. The hierarchical network structure of DBN can be constructed by building two or more pre-trained RBMs. The output patterns of hidden neurons at the $l$-th RBM can be seen as the next input at the $(l+1)$-th RBM. The conditional probability of a hidden neuron $j$ at the $l$-th RBM is defined by Eq.(\ref{eq:prob_dbn}).
%\begin{equation}
%\label{eq:prob_dbn}
%p(h_j^{l} = 1 | \bvec{h}^{l-1})= sigmoid(c^{l}_j + \sum_{i}W^{l}_{ij} h^{l-1}_{i}),
%\end{equation}
%where $c^{l}_j$ and $W^{l}_{ij}$ are the parameters for a hidden neuron $j$ and the weight at the $l$-th RBM, respectively. $\bvec{h}^{0} = \bvec{v}$ is the given input data. When the trained DBN takes a supervised learning for a classification task, an output layer is added to the last layer, then the output probability $y_k$ for a category $k$ is calculated by Softmax at the output layer as shown in Eq.(\ref{eq:softmax}). 
%\begin{equation}
%\label{eq:softmax}
%y_k = \frac{\exp(z_{k})}{\sum^{M}_{j} \exp(z_j)},
%\end{equation}
%where $z_{j}$ is an output pattern of a hidden neuron $j$ at the output layer. $M$ is the number of output neurons. The error between the output $y_k$ and the teacher signal is minimized. 

\subsection{Neuron Generation and Annihilation Algorithm of RBM}
\label{subsec:adaptive_rbm}
\begin{figure}[tbp]
\begin{center}
\subfigure[Neuron generation]{\includegraphics[scale=0.5]{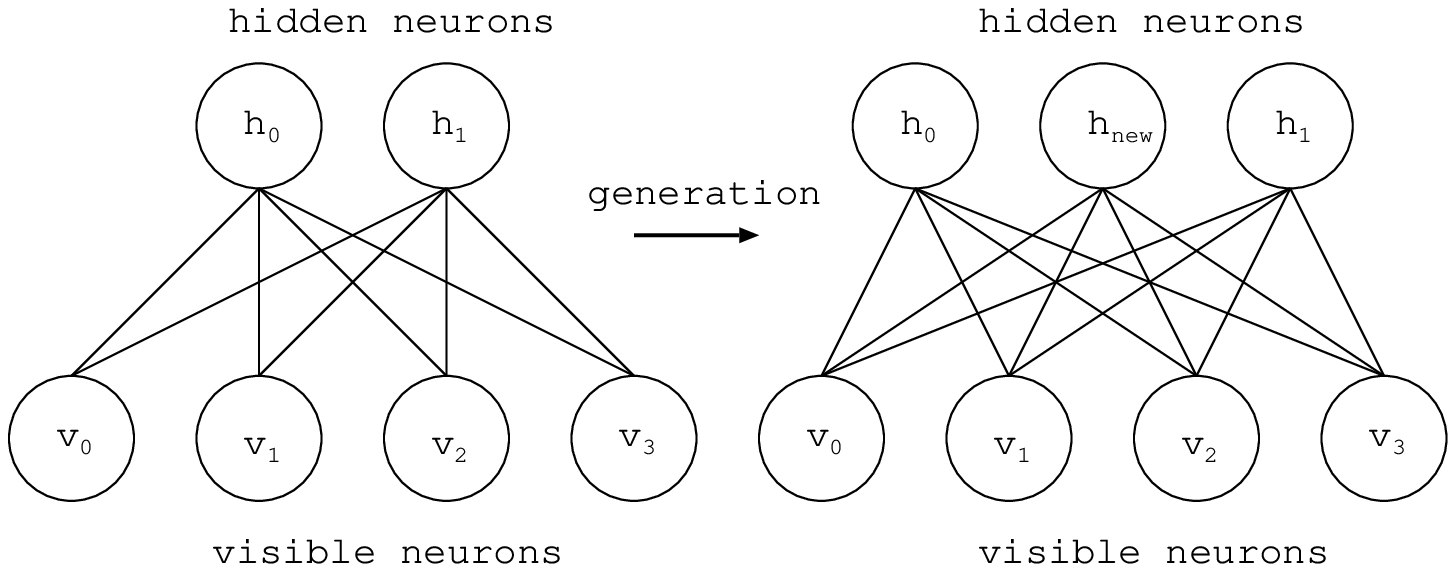}\label{fig:neuron_generation}}
\subfigure[Neuron annihilation]{\includegraphics[scale=0.5]{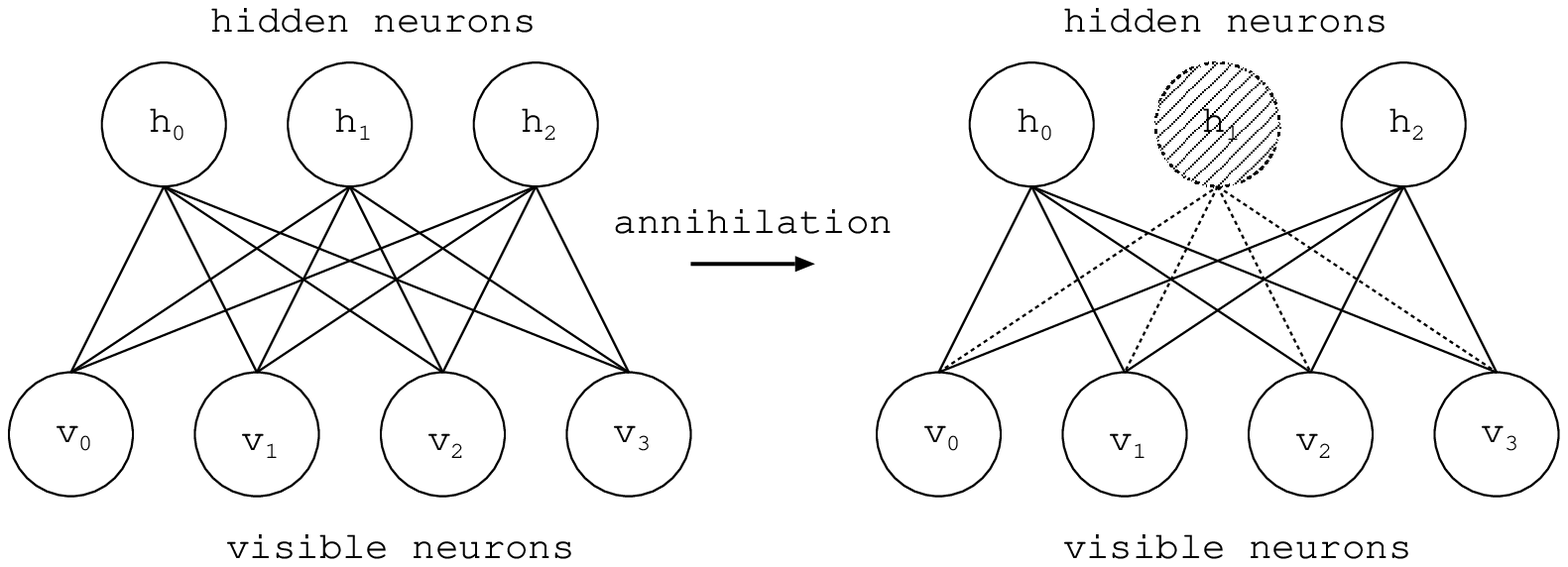}\label{fig:neuron_annihilation}}
%\vspace{-3mm}
\caption{Adaptive RBM}
\label{fig:adaptive_rbm}
\end{center}
\end{figure}

Generally, the decision of the optimal structure to input data depends on the skill of the network designer, since many trials for finding the optimal set of parameters should be required to reach higher classification. In order to make a solution, the adaptive structural learning method of RBM (Adaptive RBM) has been proposed. The method has an outstanding function of determination for the structure of RBM \cite{Hinton06,Hinton12} which has the self-organized algorithm by hidden neuron generation and deletion during learning phase. The method as shown in Fig.\ref{fig:adaptive_rbm} gives a good solution for the problem that the traditional RBM has a stationary network structure during training even if the network is not enough hidden neurons to classify input data. The Adaptive RBM can determine the suitable number of hidden neurons by observation the variance of weights and its coefficients during the training.

The algorithm monitors the training situation of the network with the fluctuation of weight vector, the Walking Distance (WD) \cite{Ichimura04}. WD is defined as the difference between the past variance and the current variance for learning parameters such as weight during training. The basic idea of \cite{Ichimura04} is as follows. If the network does not have enough neurons to classify them sufficiently, then WD will tend to fluctuate large after the long training process. The situation shows that some hidden neurons may not represent an ambiguous pattern due to the lack of the number of hidden neurons. In order to represent ambiguous patterns into two neurons, a new neuron is inserted to inherit the attributes of the parent hidden neuron as shown in Fig.~\ref{fig:neuron_generation}.

Adaptive RBM employs the neuron generation with inner product of the variance of monitoring two kinds of parameters $\bvec{c}$ and $\bvec{W}$ except $\bvec{b}$. The reason for the exclusion of $\bvec{b}$ is that the parameter is observed the oscillation according to the input patterns, because the input signals will include some noise data. We showed the detailed algorithms, equations, and experiment results in the paper \cite{Kamada18_Springer}.

On the other hand, after neuron generation process, we can see that the network has some unnecessary or redundant neurons.  Since these neurons are not contributed to infer for the input data set, the network takes much computational cost. The neuron annihilation algorithm can remove the specified neurons due to output activation of signal. Fig.~\ref{fig:neuron_annihilation} shows that the corresponding neuron is annihilated.

\subsection{Layer Generation Algorithm of DBN}
\label{subsec:adaptive_dbn}
By using the idea of Adaptive RBM, the adaptive structural learning method of DBN (Adaptive DBN) was proposed. The method has the self-organized algorithm by hidden neuron generation and deletion during learning phase. The number of RBM layers is also automatically defined by the layer generation method \cite{Kamada16_SMC, Kamada16_ICONIP, Kamada16_TENCON}, where each RBM is followed by Adaptive RBM method in section \ref{subsec:adaptive_rbm}. DBN has data representation power that performs the specified features from an abstract concept to concrete representation at each layer in the direction to output layer. Adaptive DBN can automatically adjust an optimal network structure by the self-organization.

\begin{figure*}[tbp]
\centering
\includegraphics[scale=0.8]{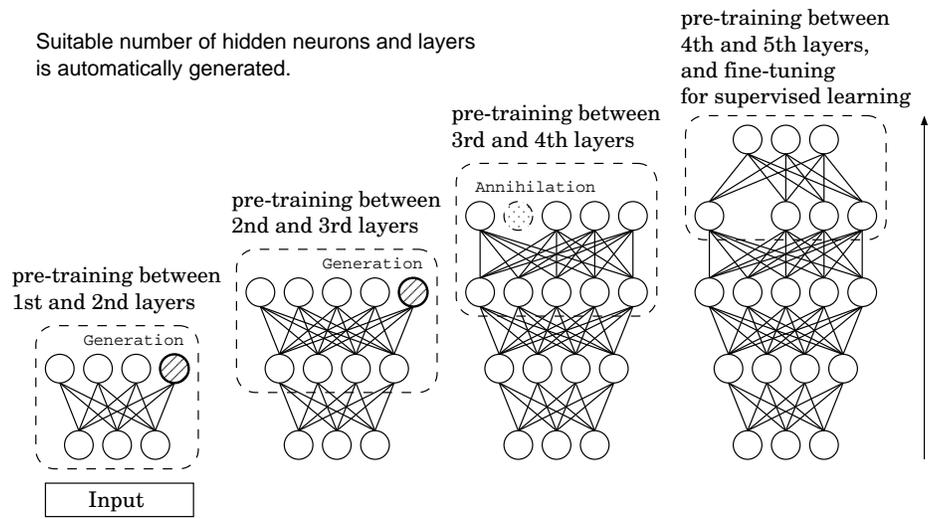}
\caption{Overview of Adaptive DBN}
\label{fig:adaptive_dbn}
\end{figure*}

The WD and the energy function at each RBM layer during learning process of Adaptive DBN was observed. If the values of both WD and the energy function do not become small values, then a new RBM will be generated to keep the suitable network structure for the data set, since the RBM has lacked data representation capability to figure out an image of input patterns. Therefore, the condition for layer generation is defined by using the total WD and the energy function. Fig.~\ref{fig:adaptive_dbn} shows the overview of layer generation in Adaptive DBN. Please see the \cite{Kamada18_Springer} for details.

\section{Object detection algorithm}
\label{sec:dbn_detection}
Image recognition methods are mainly classified into three tasks, classification, object detection, and segmentation. Images are usually complex and contain multiple objects. The assign of a label with image classification models can become uncertain because the classification models cannot detect semantic objects for localization. Therefore, the object detection models are more appropriate to identify multiple relevant objects in a single image. In this paper, we focus on a new object detection method by using Adaptive DBN.

\subsection{Object detection by CNN}
\label{subsec:cnn_detection}
There are many object detection algorithms which find semantic objects in images. Open CV, which is one of the most famous computer vision library, provides some function for extracting contours from images or machine learning algorithms for object detection such as facial detection \cite{OpenCV}.

In deep learning, R-CNN \cite{RCNN}, YOLO \cite{YOLO}, and SSD \cite{SSD} are famous object detection algorithms for finding B-Box in images. R-CNN provides a simple detection algorithm by using the training result of CNN. The method first extracts a candidate small region from an image and then calculates output values for the extracted region in trained network. The method decides that the region includes a semantic object (B-Box) if the output value is higher than the pre-determined threshold. YOLO \cite{YOLO} and SSD \cite{SSD} are extension methods of R-CNN with respect to finding the optimal size of regions. The basic idea of these methods is to split an image to fixed size grids, and then the optimal candidate region is determined by repeating adjustment of the grid size.

While these methods are used to detect B-Boxes in images, Xiaosong et.al proposed the generation method which estimates probability of semantic objects for given image as a heatmap \cite{CXR8}. The generated heatmap is also available for segmentation of images since the heatmap represents the likelihood map of pathologies. The method utilizes that convolution and pooling layers in CNN form two dimensional array. A heatmap is generated by product of activation values of last pooling layer and weights of next prediction layer (full connected layer) for given input. Although max pooling is often used in pooling layer of CNN, the activation values of the last pooling layer is calculated by Log-Sum-Exp (LSE) for representation of heatmap.

\subsection{Object detection by DBN}
\label{subsec:dbn_detection}
%In this paper, a new object detection algorithm is proposed by using the training result of the Adaptive DBN. We applied two important ideas of object detection in CNN to DBN. One is development of the model with high classification accuracy. The other is how to find optimal candidate regions. In previous research, our proposed Adaptive DBN showed high classification accuracy for several image benchmark data sets. Therefore, the basic idea is similar to famous object detection methods of YOLO \cite{YOLO}, and SSD \cite{SSD}, but the method of CNN cannot be applied to Adaptive DBN architecture because of the image feature of convolutional filter. Our proposed method divides the image into $N$ regions by Voronoi diagram and uses the output probability of each category for $N$ regions. {\bf Algorithm \ref{alg:dbn_detection}} shows the detailed algorithm of our proposed detection method.

In our previous research, the Adaptive DBN as described in section \ref{sec:adaptive_dbn} was applied to classification task. Since the method showed high classification accuracy for several image benchmark data sets, the method can depict the signal flow in the trained DBN network for given input by pursuing the network path of DBN from input neurons to output neurons \cite{Kamada18_AIKE}. The method is an explicit knowledge acquisition method and can reach better classification accuracy by the partial modification of path and weights. On the contrary, this paper proposes a novel method for object detection task by using the signal flow and weights of the trained DBN network.

We applied the basic ideas of object detection in CNN such as R-CNN, YOLO, and SSD to DBN. However, the methods of CNN cannot be applied to DBN architecture because of the image feature of convolutional filter. Our proposed method divides the image into $N$ regions by Voronoi diagram and uses the output probability of each category for $N$ regions. {\bf Algorithm \ref{alg:dbn_detection}} shows the detailed algorithm of our proposed detection method.

\begin{figure*}[tbp]
\centering
\includegraphics[scale=0.6]{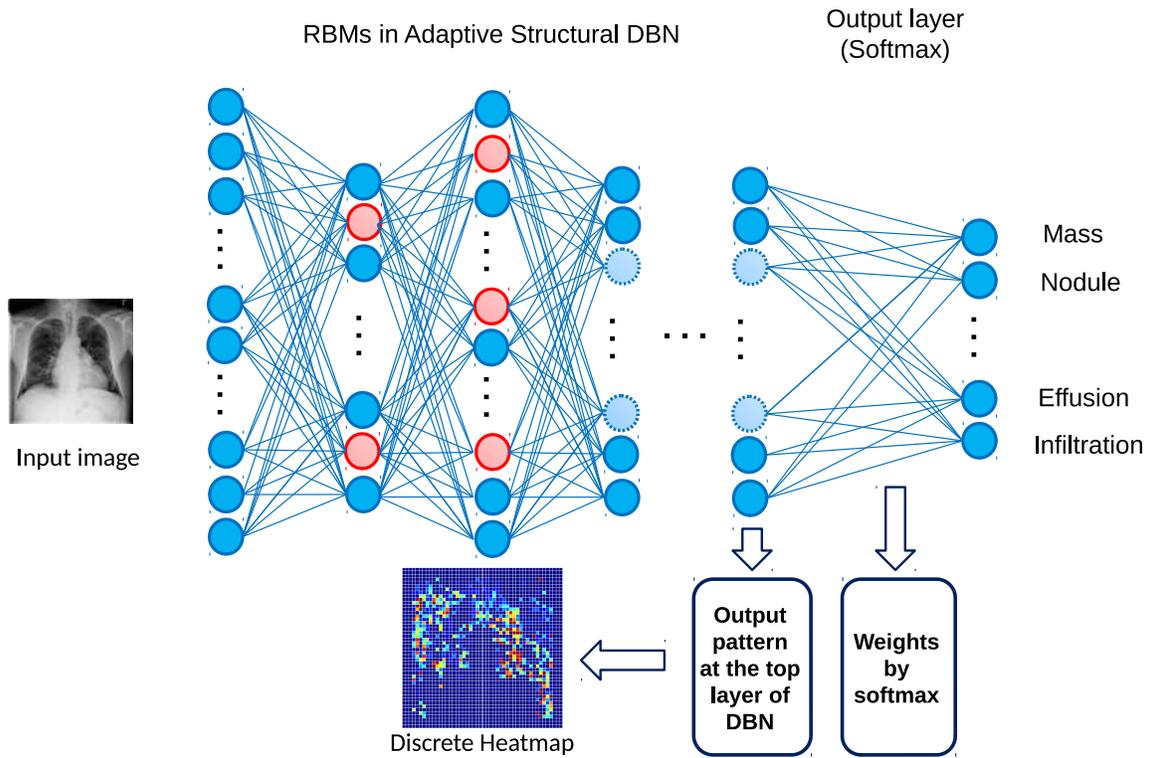}
\caption{An overview of heatmap}
\label{fig:heatmap_overview}
\end{figure*}

Moreover, the generation method of heatmap is implemented on Adaptive DBN in addition to detection of B-Box. The CNN \cite{CXR8} can calculate a heatmap by product of activation values of pooling layer and weights of prediction layer for given input because the pooling layer forms two dimensional array. However, hidden neurons in DBN are represented by one dimensional array. In DBN, the product of activation values of last hidden layer and weights of next prediction layer (softmax) is represented by one dimensional feature vector. It means there are no mapping information between the vector and input image. Therefore, we solved the problem by backward calculation from the vector in output layer to input layer. Fig.~\ref{fig:heatmap_overview} shows an overview of calculation procedure of a heatmap. In this paper, the value of a heatmap has the range $[0, 255]$ and it is represented by jet color array \cite{colormap_jet} as same as the paper \cite{CXR8}.

\begin{algorithm}[tbp]
\caption{Detection algorithm of Adaptive DBN}
\label{alg:dbn_detection}                        
\begin{algorithmic}[1]
\STATE Let an input be $w \times h$ pixel image.
\STATE Divide the image into $N$ regions $V = \{v_1, \cdots, v_i, \cdots, v_N\}$. $v_i$ is a region. In this paper, each region $v_i$ is determined by Voronoi diagram as Eq.~(\ref{eq:voronoi}).
\begin{equation}
v_i = \{p | d(p, p_j) \leq d(p, p_k), k \neq j\}
\label{eq:voronoi}
\end{equation}
where $d()$ is a distance function. In this paper, the points are randomly determined.
\STATE Each region $v_i$ is given to the trained DBN network and then output probability of each category (output) is calculated. If the probability of a class is more than pre-determined threshold $T_{1} (0 < T_{1} < 1)$, the system decide to the region is candidate region for detection.
\STATE Multiple candidate images are generated by slightly adjusting the size of each decided candidate region $v_{i}$. The candidate image has the size $n \times m$ and its center coordinate is same as $v_i$. $n$ and $m$ are selected in $N = \{N_{min}, \cdots, n, \cdots, N_{max}\}$ and $M = \{M_{min}, \cdots, m, \cdots, M_{max}\}$, respectively.
\STATE The generated candidate images are given to the trained DBN network and then output probability of each category (output) is calculated. If the probability of a class is more than pre-determined threshold $T_{2} (0 < T_{1} < T_{2} < 1)$, the system decide to the region is B-Box for the class.
\end{algorithmic}
\end{algorithm}

\section{Experimental Results}
\label{sec:exe}
In this paper, the effectiveness of our proposed method for CXR8 image benchmark data set is verified. The classification and detection performance are compared with some CNN methods.

\subsection{ChestX-ray8}
\label{subsec:exe_cxr8}
\begin{figure}[tbp]
  \centering
  %\subfigure[No Finding]{\includegraphics[scale=0.35]{fig/No_Finding.eps}\label{fig:cxr8_sample_no_finding}}
  \subfigure[Atelectasis]{\includegraphics[scale=0.4]{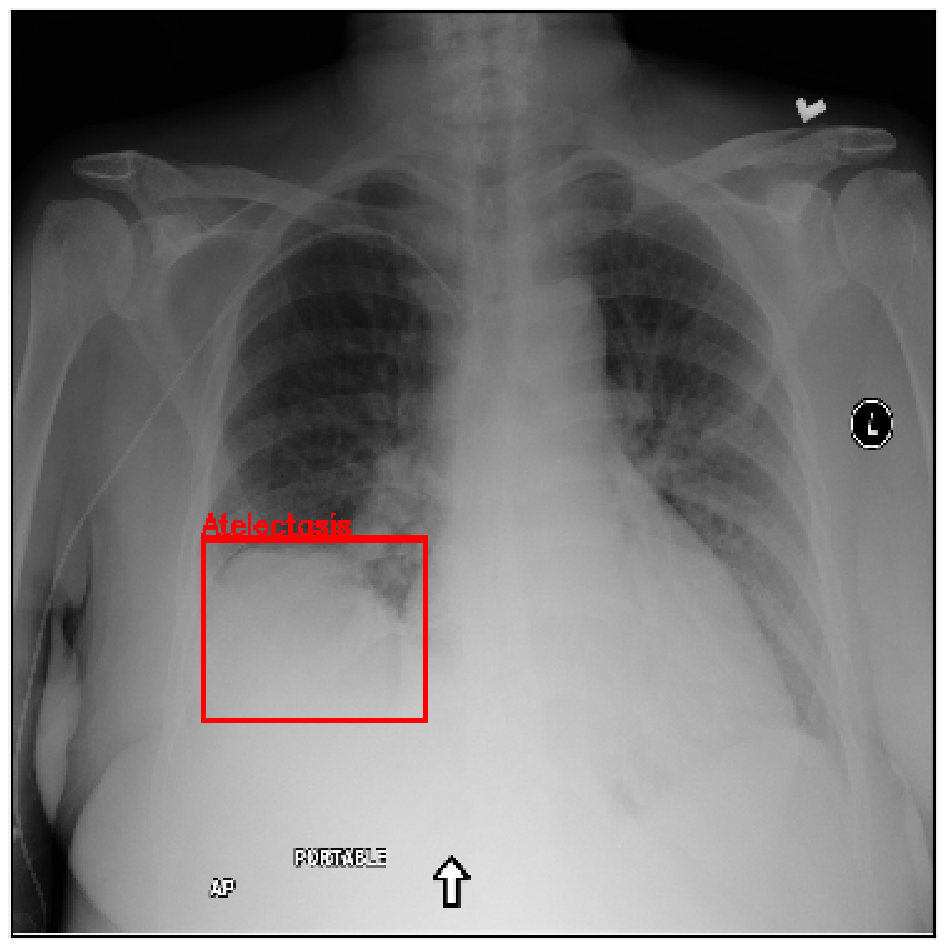}\label{fig:cxr8_sample_atelectasis}}
  %\subfigure[Cardiomegaly]{\includegraphics[scale=0.5]{fig/Cardiomegaly.eps}\label{fig:cxr8_sample_cardiomegaly}}
  %\subfigure[Effusion]{\includegraphics[scale=0.5]{fig/Effusion.eps}\label{fig:cxr8_sample_effusion}}
  %\subfigure[Infiltration]{\includegraphics[scale=0.5]{fig/Infiltration.eps}\label{fig:cxr8_sample_infiltration}}
  %\subfigure[Pneumonia]{\includegraphics[scale=0.5]{fig/Pneumonia.eps}\label{fig:cxr8_sample_pneumonia}}
  %\subfigure[Pneumothorax]{\includegraphics[scale=0.5]{fig/Pneumothorax.eps}\label{fig:cxr8_sample_pneumothorax}}
  \subfigure[Mass]{\includegraphics[scale=0.4]{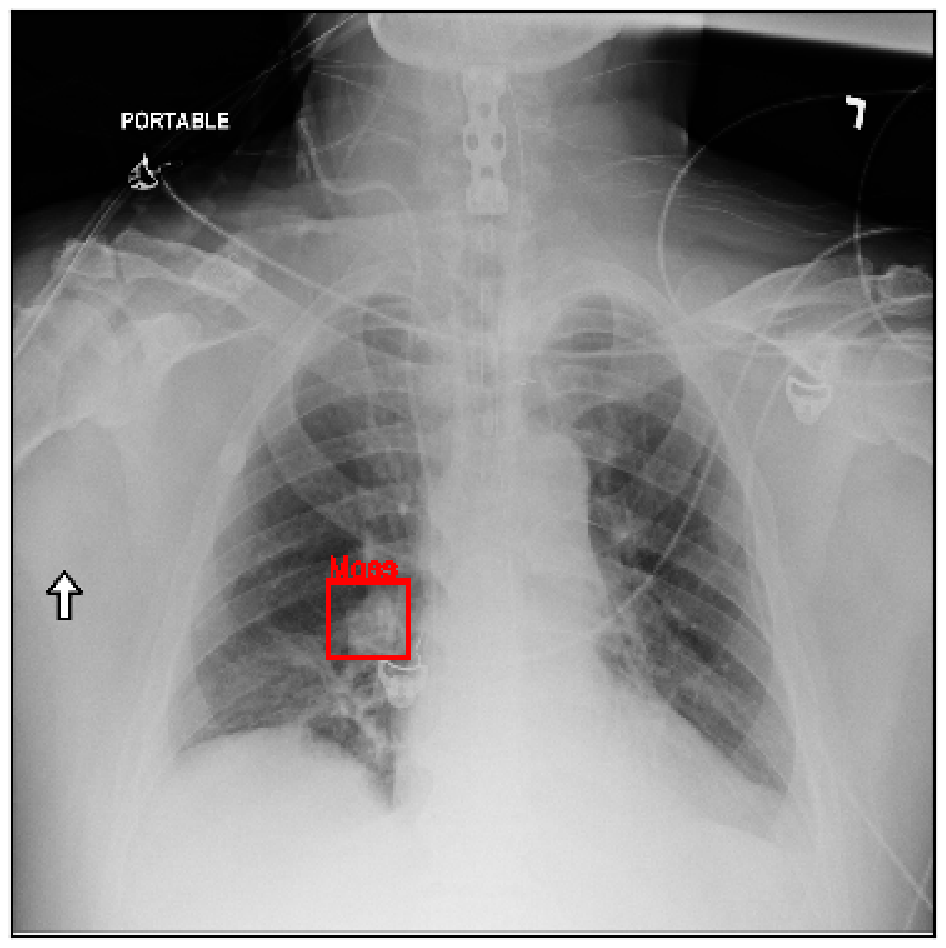}\label{fig:cxr8_sample_mass}}
  %\subfigure[Nodule]{\includegraphics[scale=0.5]{fig/Nodule.eps}\label{fig:cxr8_sample_nodule}}
  \caption{Image examples of CXR8}
  \label{fig:cxr8_sample}
\end{figure}

CXR8 \cite{CXR8} is one of the most commonly accessible radio-logical examination for many lung diseases. The data consists of 112,120 images collected by 30,805 patients. As shown in Table \ref{tab:data_cxr_category}, nine class labels of normal state and eight diseases including cancer are defined for classification. The data is divided into training set and test set, and the classification accuracy and ROC curve with several well-known deep network such as VGG16, GoogLeNet, and ResNet, are reported on the original paper \cite{CXR8}. In addition, 984 Bounding Boxes (B-Box) are provided for localization. Fig.~\ref{fig:cxr8_sample} shows image examples of CXR8. The red rectangle in the image shows given B-Box.

\begin{table}[tbp]
\caption{Categories and data size of CXR8}
\vspace{-3mm}
\label{tab:data_cxr_category}
\begin{center}
%\scalebox{0.8}[0.8]{
\begin{tabular}{l|r|r}
\hline \hline
\multicolumn{1}{c|}{Category} & Classification & Detection (B-Box) \\ \hline
\hline
No Finding & 60,361 & -\\ \hline
Mass & 5,782 & 85\\ \hline
Nodule & 6,331 & 79 \\  \hline
Atelectasis & 11,559 & 180\\  \hline
Cardiomegaly & 2,776 & 146 \\ \hline
Effusion & 13,317 & 153  \\  \hline
Infiltration & 19,894 & 123\\ \hline
Pneumonia & 1,431 & 120 \\  \hline
Pneumothorax & 5,302 & 98\\ \hline
\hline
Total & 112,120 & 984\\ \hline
\hline
\end{tabular}
%} 
\end{center}
%\vspace{-5mm}
\end{table}

\subsection{Classification results}
\label{subsec:exe_classification}
In this subsection, the effectiveness of proposed Adaptive DBN for eight diseases of CXR8 is verified. The parameters for Adaptive DBN are as follows. The training algorithm was Stochastic Gradient Descent (SGD) method, the batch size was 100, the learning rate was 0.005, the initial number of hidden neurons was 400, $\theta_G = 0.001, \theta_A = 0.100$, $\theta_{L1} = 0.05$, and $\theta_{L2} = 0.05$. We used the computer with the following specifications during training: CPU: Intel(R) 24 Core Xeon E5-2670 v3 2.3GHz, GPU: Tesla K80 4992 24GB $\times$ 3, Memory: 64GB, OS: Cent OS 6.7 64 bit. The classification accuracy of the Adaptive DBN for test data was compared with the several CNN methods, which are reported on the paper \cite{CXR8}.

Table~\ref{tab:result_cxr8_classification} shows the classification accuracy of test data for each diseases in CXR8. `GoogLeNet', `VGG16', and `ResNet-50' are the result of transfer learning with well-known CNN methods \cite{CXR8}. The classification accuracy of ResNet-50 was 81.4\% which was higher value than the other CNN methods. On the other hand, the classification accuracy of the Adaptive DBN was more than 90\% for all diseases and it achieved the highest value among the comparison methods.

\begin{figure}[tbp]
\centering
\includegraphics[scale=0.3]{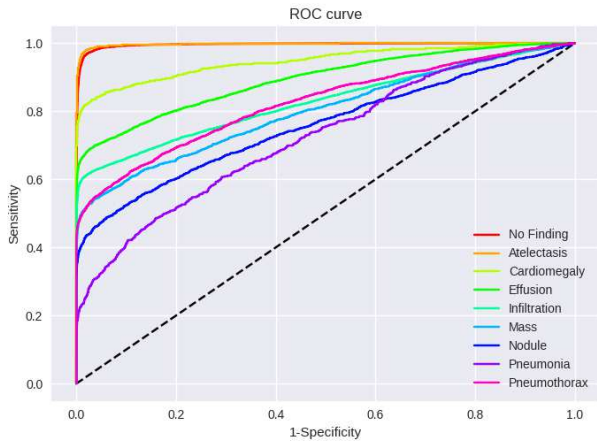}
\caption{ROC curve}
\label{fig:result_cxr8_roc}
\end{figure}

Fig.~\ref{fig:result_cxr8_roc} shows the ROC curve of the Adaptive DBN. The ROC curve plots true positive rate such as sensitivity on vertical axis and false positive rate (1-specificity) on horizontal axis. The Adaptive DBN showed better performance since the area under the Adaptive DBN curve was larger than the other CNN methods (Please see the paper \cite{CXR8} for details). 

\begin{table}[tbp]
\caption{Classification accuracy for test data in CXR8}
\label{tab:result_cxr8_classification}
\centering
\begin{tabular}{c||r|r|r|r}
\hline \hline
     & \multicolumn{4}{c}{Method}  \\ \cline{2-5}
Category  & GoogLeNet & VGG16 & ResNet-50 & Adaptive DBN   \\ \hline
\hline
No Finding    - & - & - & - & 90.0\%  \\ \hline
Mass          & 54.6\% & 51.0\% & 56.0\% & 96.3\% \\ \hline
Nodule        & 55.7\% & 65.5\% & 71.6\% & 97.2\% \\ \hline
Atelectasis   & 63.0\% & 62.8\% & 70.6\% & 94.5\% \\ \hline
Cardiomegaly  & 70.5\% & 70.8\% & 81.4\% & 98.1\% \\ \hline
Effusion      & 68.7\% & 65.0\% & 73.6\% & 97.2\% \\ \hline
Infiltration  & 60.8\% & 58.9\% & 61.2\% & 96.0\% \\ \hline
Pneumonia     & 59.9\% & 51.0\% & 63.3\% & 99.9\% \\ \hline
Pneumothora   & 78.2\% & 75.1\% & 78.9\% & 98.1\% \\ \hline
\hline
\end{tabular}
\end{table}

\subsection{Detection results}
\label{subsec:exe_detection}
By using the training result of the Adaptive DBN in section \ref{subsec:exe_classification}, the proposed detection algorithm ({\bf Algorithm \ref{alg:dbn_detection}}) was applied to 984 images with given B-Box. In the simulation results, the following parameters were used for the detection algorithm; $T_{1} = 0.5$, $T_{2} = \{0.7, 0.9\}$.

Table.~\ref{tab:result_cxr8_detection} shows the detection accuracy. `ResNet-50' is the result of transfer learning described in \cite{CXR8} as same as Table~\ref{tab:result_cxr8_classification}. In this paper, the standard Intersection over Union ratio (IoU) was used to examine the accuracy. This is, the detected B-Box is decided to be correct if the ratio of intersection between given B-Box and detected B-Box is more than the pre-determined threshold. We set this threshold to 75\% as same as the original paper. The detection accuracy of our proposed method is higher than the CNN method for all diseases. Especially, our method was able to detect all diseases with more 90\% ratio when $T_{2}$ was $0.5$.

\begin{table}[tbp]
\caption{Detection accuracy}
\vspace{-3mm}
\label{tab:result_cxr8_detection}
\begin{center}
%\scalebox{0.8}[0.8]{
\begin{tabular}{c|r|r|r}
\hline \hline
% \multicolumn{3}{c|}{} & \multicolumn{4}{c|}{Accuracy}  \\ \cline{4-6}
& & \multicolumn{2}{c}{Adaptive DBN}  \\ \cline{3-4}
Category & ResNet50 \cite{CXR8} & $T_{2} = 0.7$ & $T_{2} = 0.9$  \\ \hline
\hline
Atelectasis & 47.2\% & 78.9\% & 86.1\% \\ \hline 
Cardiomegaly & 68.4\% & 99.3\% & 100.0\% \\ \hline 
Effusion & 45.0\% & 85.6\% & 92.8\% \\ \hline 
Infiltration & 47.9\% & 91.9\% & 93.5\% \\ \hline 
Pneumonia & 35.0\% & 84.2\% & 92.5\% \\ \hline 
Pneumothorax & 23.4\% & 80.6\% & 83.7\% \\ \hline 
Mass & 25.8\% & 88.2\% & 91.8\% \\ \hline 
Nodule & 5.0\% & 72.2\% & 77.2\% \\ \hline 
\hline
Total & - & 85.7\% & 90.4\% \\ \hline
\hline
\end{tabular}
%} 
\end{center}
%\vspace{-5mm}
\end{table}

\subsection{Investigation of the generated heatmap}
\label{subsec:exe_heatmap}
\begin{figure}[!t]
  \centering
  \subfigure[Original image]{\includegraphics[scale=0.4]{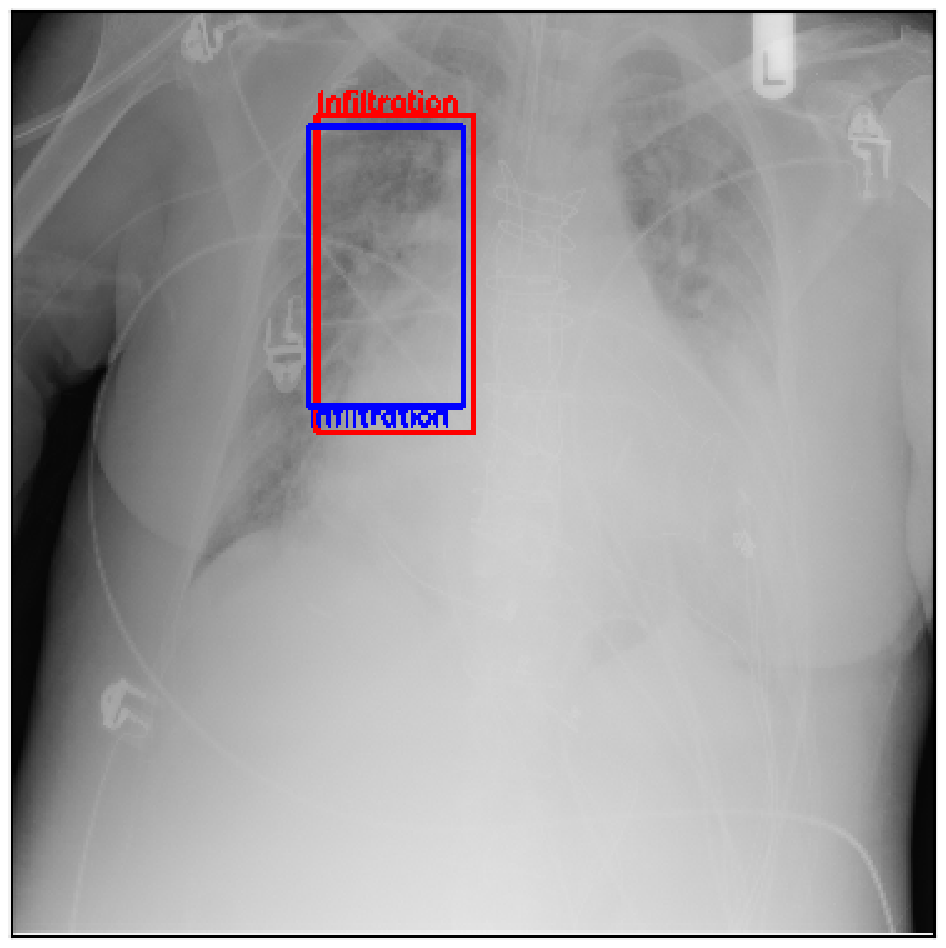}\label{fig:case1-o}}
  \subfigure[Heatmap]{\includegraphics[scale=0.4]{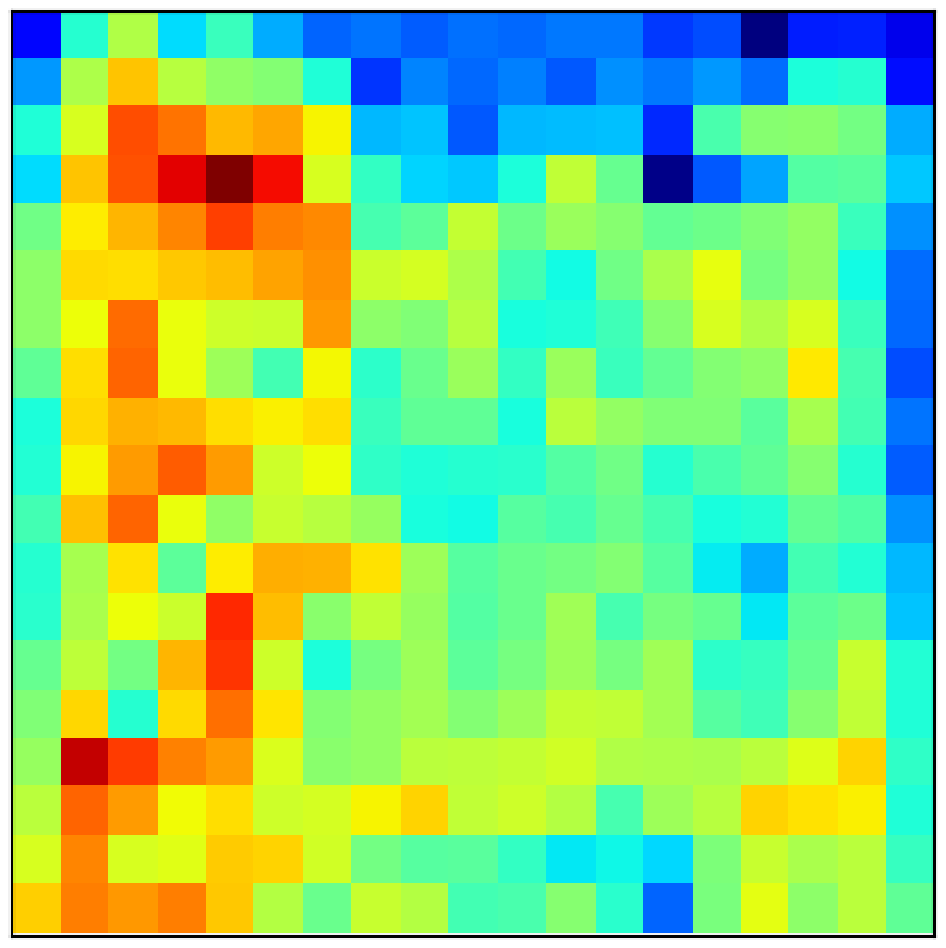}\label{fig:case1-h}}
  \caption{Case 1}
  \label{fig:heatmap_case1}
  %\vspace{-5mm}
%\end{figure}

%\begin{figure}[tbp]
%  \centering
  \subfigure[Original image]{\includegraphics[scale=0.4]{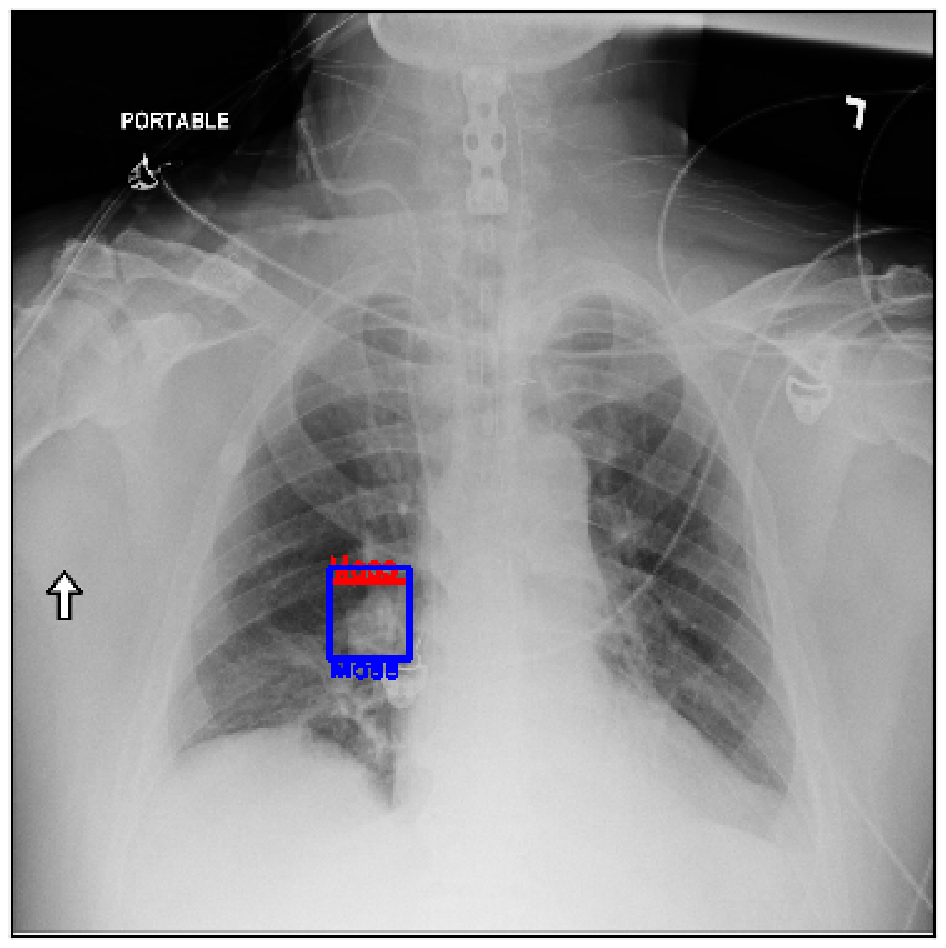}\label{fig:case2-o}}
  \subfigure[Heatmap]{\includegraphics[scale=0.4]{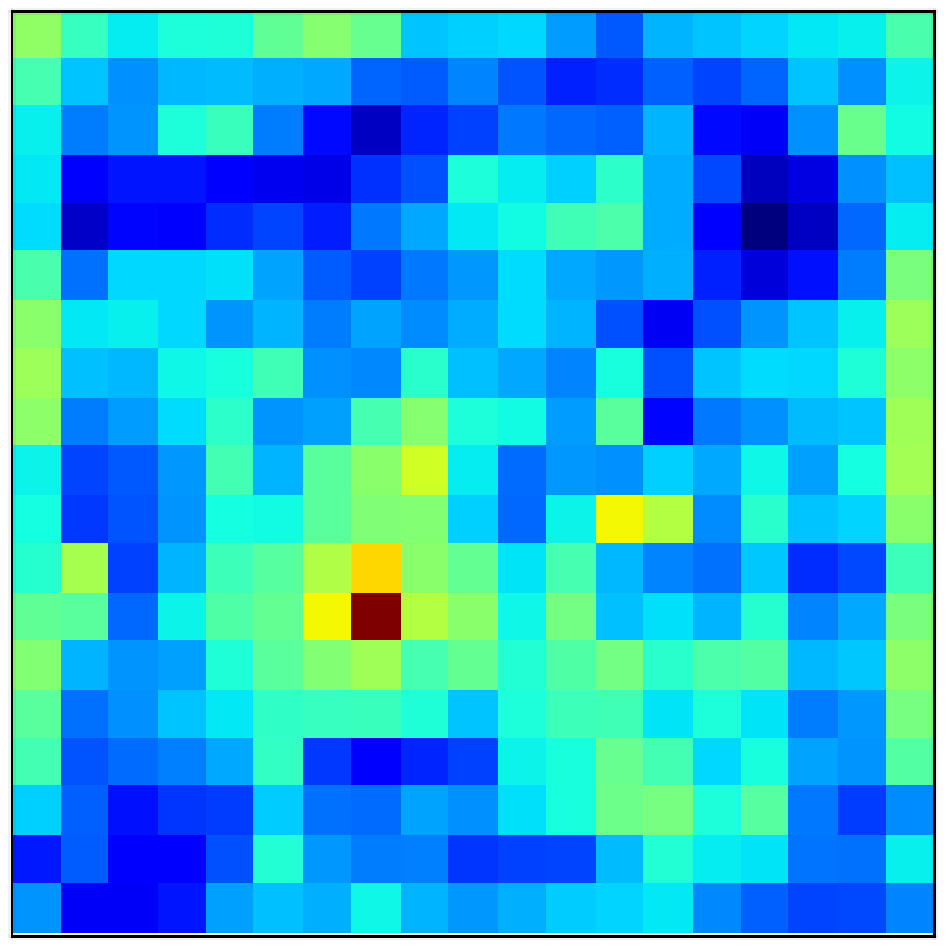}\label{fig:case2-h}}
  \caption{Case 2}
  \label{fig:heatmap_case2}
  %\vspace{-5mm}
%\end{figure}

%\begin{figure}[tbp]
%  \centering
  \subfigure[Original image]{\includegraphics[scale=0.4]{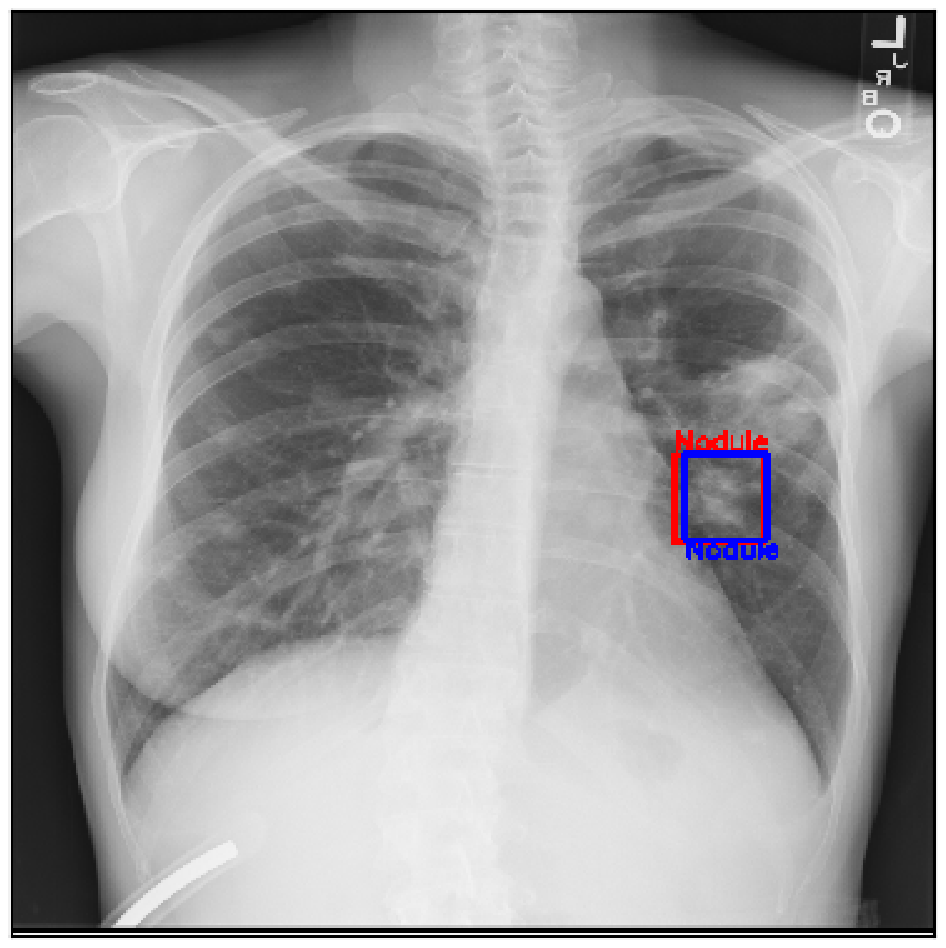}\label{fig:case3-o}}
  \subfigure[Heatmap]{\includegraphics[scale=0.4]{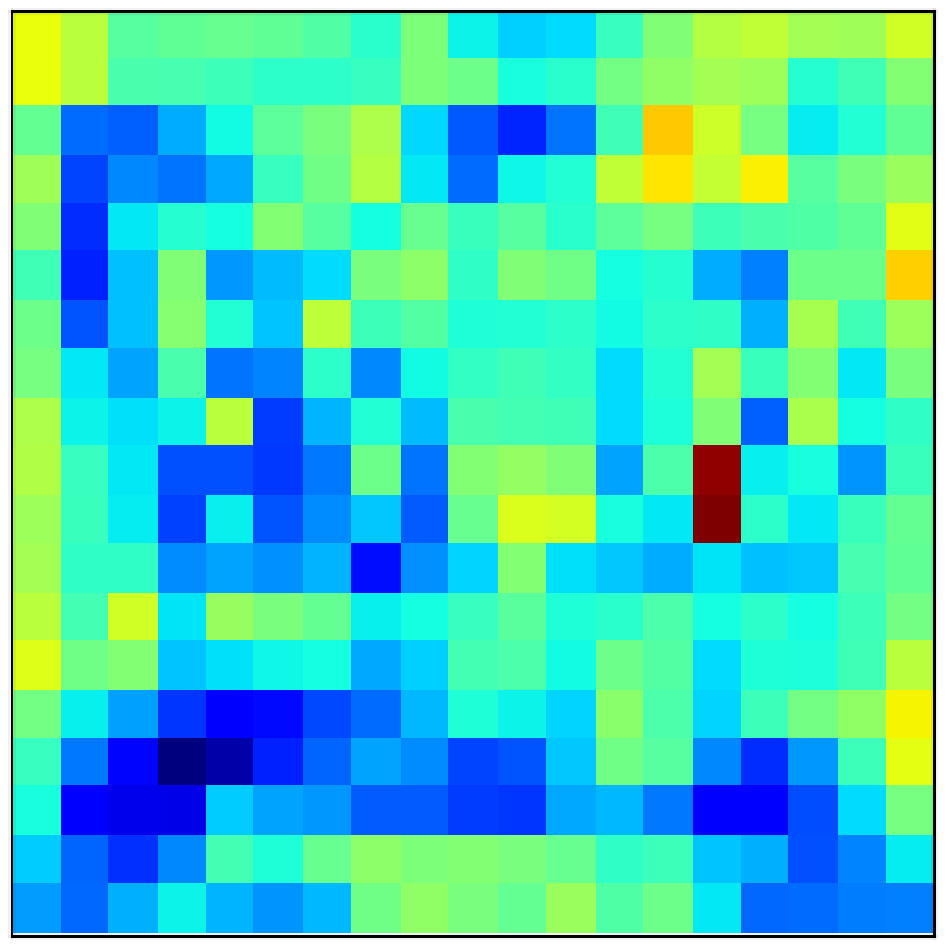}\label{fig:case3-h}}
  \caption{Case 3}
  \label{fig:heatmap_case3}
  %\vspace{-5mm}
%\end{figure}

%\begin{figure}[tbp]
%  \centering
  \subfigure[Original image]{\includegraphics[scale=0.4]{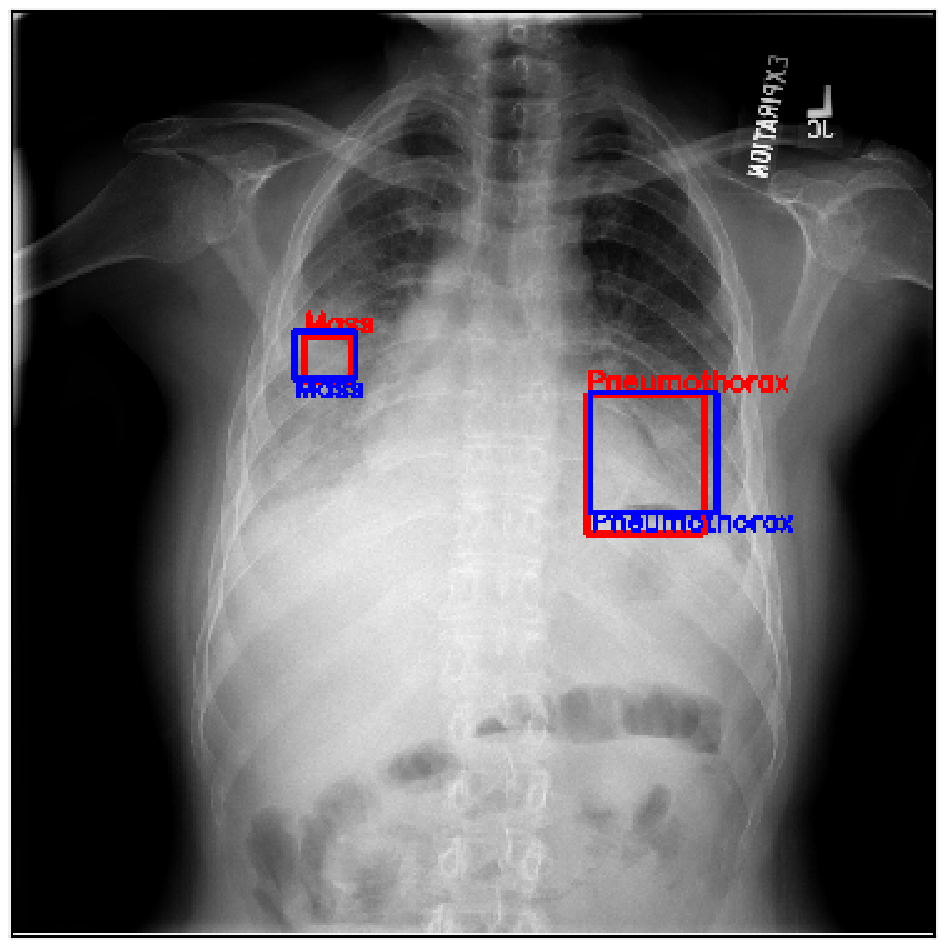}\label{fig:case4-o}}
  \subfigure[Heatmap]{\includegraphics[scale=0.4]{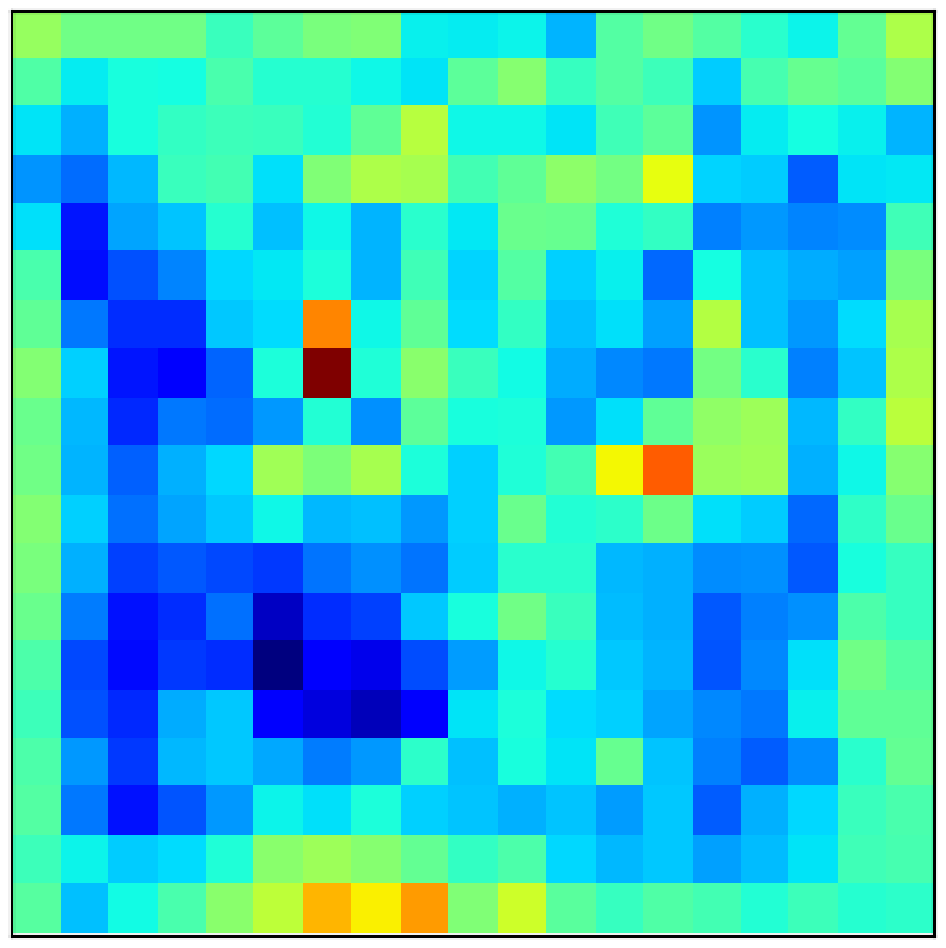}\label{fig:case4-h}}
  \caption{Case 4}
  \label{fig:heatmap_case4}
  %\vspace{-5mm}
\end{figure}

\begin{figure}[!t]
  \centering
  \subfigure[Original image]{\includegraphics[scale=0.4]{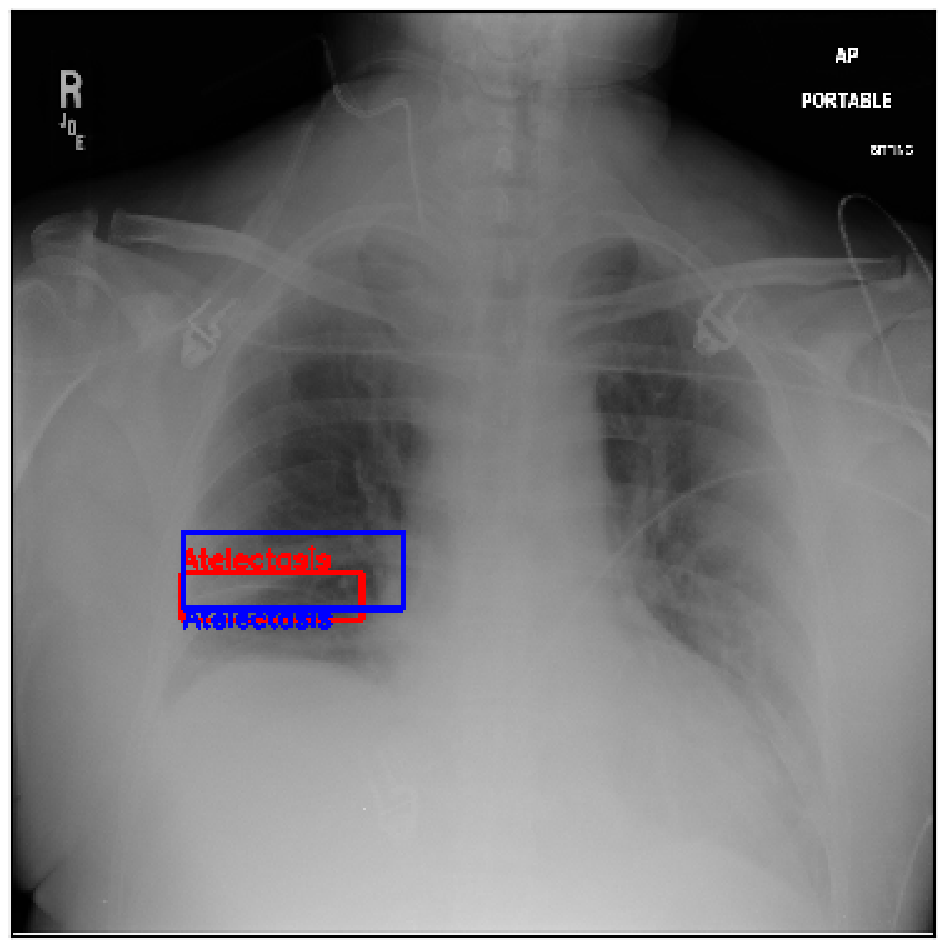}\label{fig:case5-o}}
  \subfigure[Heatmap]{\includegraphics[scale=0.4]{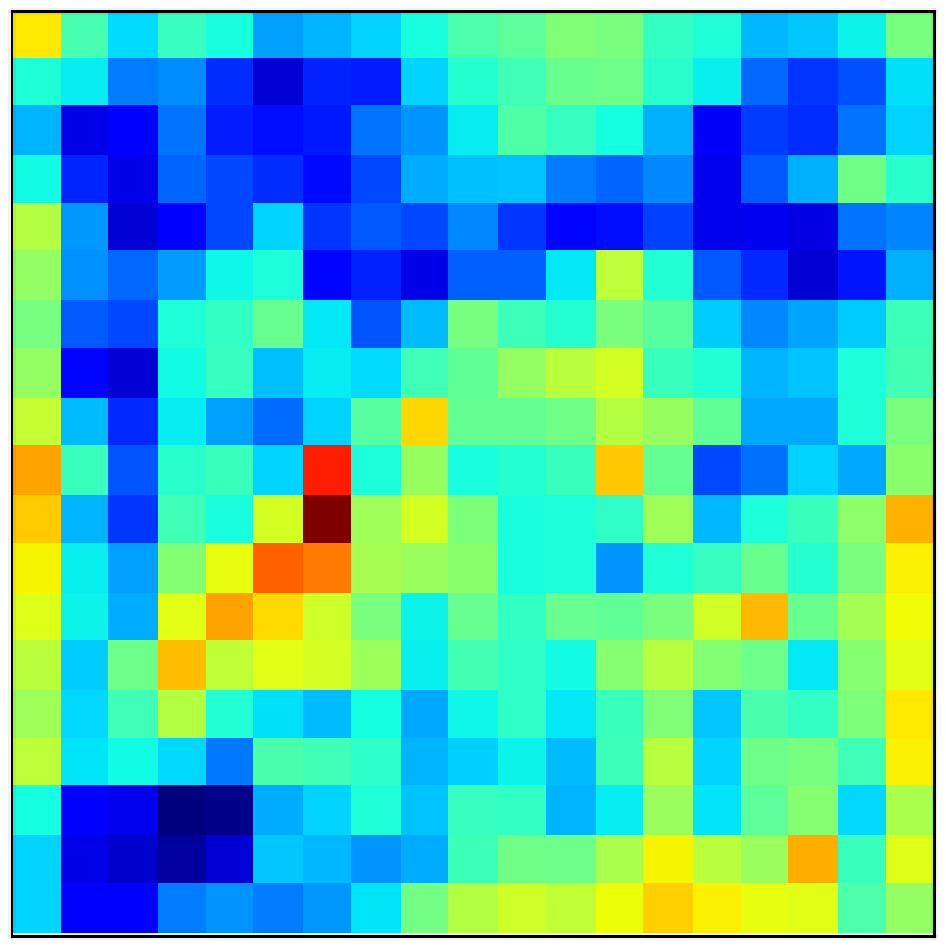}\label{fig:case5-h}}
  \caption{Case 5}
  \label{fig:heatmap_case5}
  %\vspace{-5mm}
%\end{figure}

%\begin{figure}[t]
%  \centering
  \subfigure[Original image]{\includegraphics[scale=0.4]{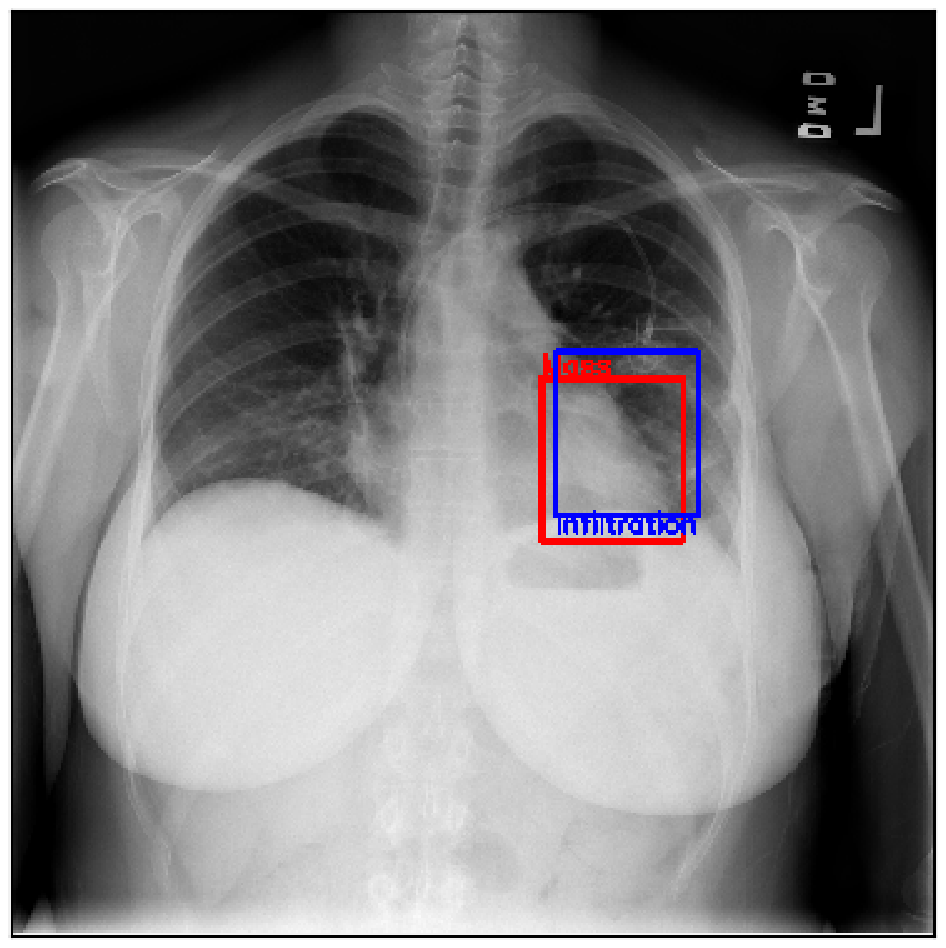}\label{fig:case6-o}}
  \subfigure[Heatmap]{\includegraphics[scale=0.4]{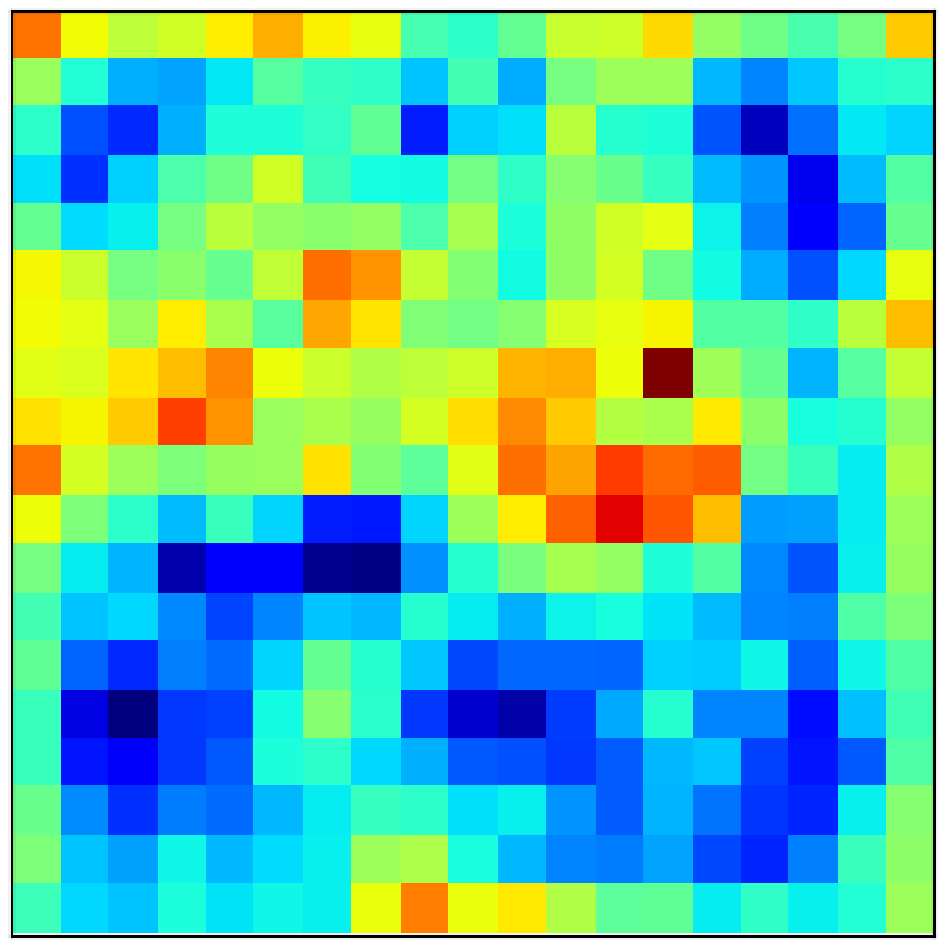}\label{fig:case6-h}}
  \caption{Case 6}
  \label{fig:heatmap_case6}
  %\vspace{-5mm}
%\end{figure}

%\begin{figure}[tbp]
%  \centering
  \subfigure[Original image]{\includegraphics[scale=0.4]{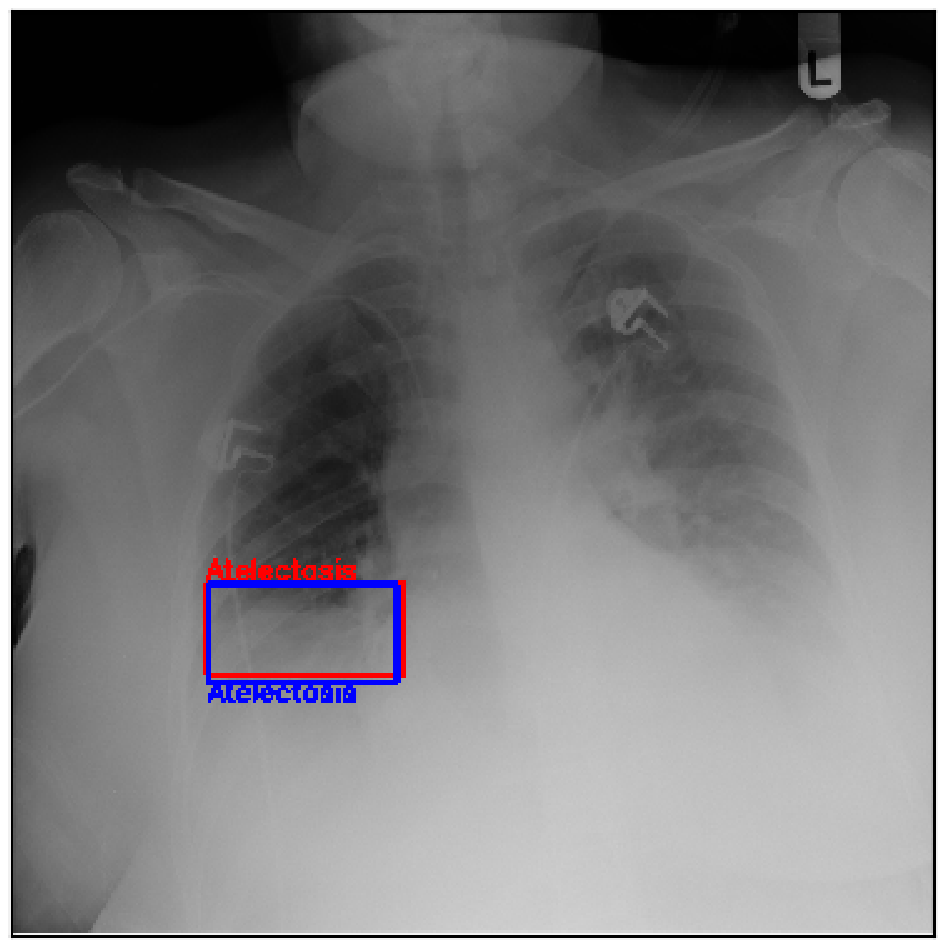}\label{fig:case7-o}}
  \subfigure[Heatmap]{\includegraphics[scale=0.4]{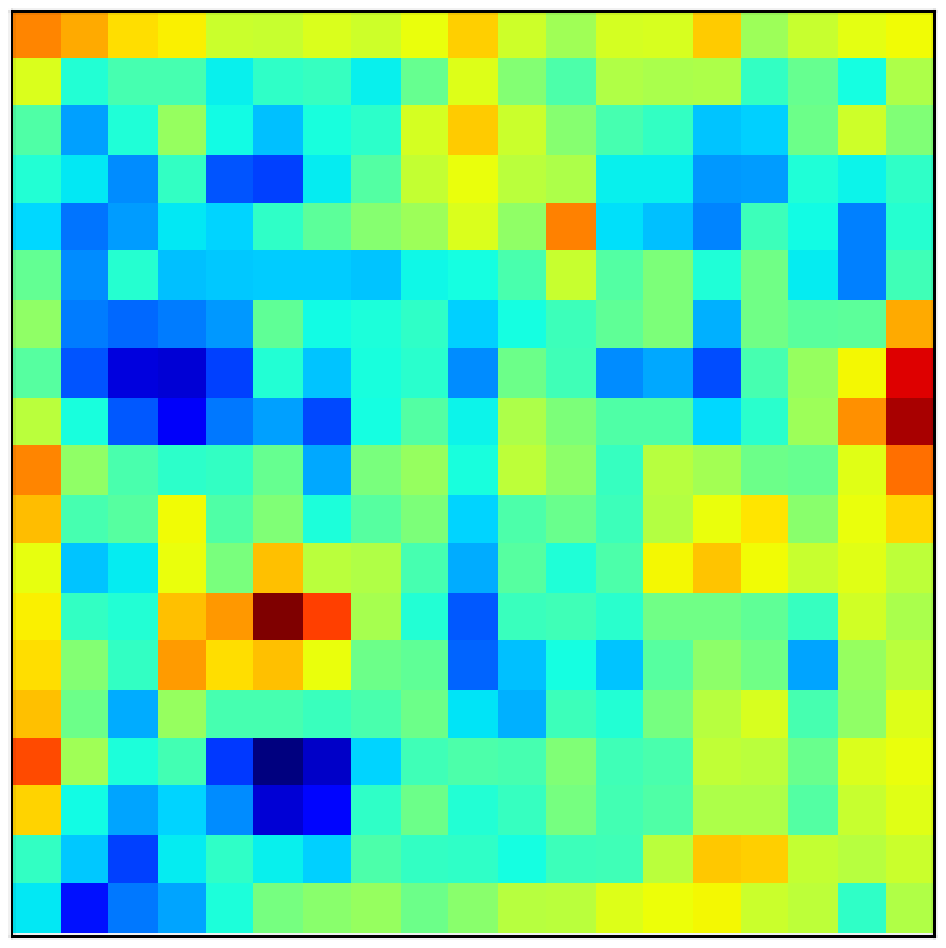}\label{fig:case7-h}}
  \caption{Case 7}
  \label{fig:heatmap_case7}
  %\vspace{-5mm}
%\end{figure}

%\begin{figure}[!t]
 % \centering
  \subfigure[Original image]{\includegraphics[scale=0.4]{00006851_033.eps}\label{fig:case1}}
  \subfigure[Heatmap]{\includegraphics[scale=0.4]{00006851_033.png-h.eps}\label{fig:case1}}
  \caption{Case 8}
  \label{fig:heatmap_case8}
  %\vspace{-5mm}
\end{figure}

By using the training result of the Adaptive DBN, the heatmap images were generated in addition to detection of B-Box. The heatmaps in Fig.~\ref{fig:heatmap_case1} to Fig.~\ref{fig:heatmap_case8} show the detection result of B-Box and the generated heatmap for some images. The red and blue rectangles in the image are given B-Box and detected B-Box, respectively. A heatmap is represented by the continuous value of range $[0, 255]$, where the color map is jet color array (red means high value, while blue means small value). The diseases for detected B-Boxes in Fig.~\ref{fig:heatmap_case1} to Fig.~\ref{fig:heatmap_case8} were as follows; Infiltration (Fig.~\ref{fig:heatmap_case1}), Mass (Fig.~\ref{fig:heatmap_case2}), Nodule (Fig.~\ref{fig:heatmap_case3}), Mass and Pneumothorax (Fig.~\ref{fig:heatmap_case4}), Atelectasis (Fig.~\ref{fig:heatmap_case5}), Infiltration (Fig.~\ref{fig:heatmap_case6}), Atelectasis (Fig.~\ref{fig:heatmap_case7}), Atelectasis (Fig.~\ref{fig:heatmap_case8}).

Overall, the red area of the generated heatmap included in both the given B-Box and detected B-Box. On the other hand, the blue or yellow areas didn't include in these B-Boxes. This tendency was seen in not only large diseases (e.g. Cardiomegaly or Infiltration) such as Fig.~\ref{fig:heatmap_case1}, but also small diseases (e.g. Mass or Nodule) such as Fig.~\ref{fig:heatmap_case2}. We consider that the experimental results caused by the discrete heatmap with binary output of final RBM layer instead of continuous heatmap. As a result, the red regions represents localization with strong relation to diseases and blue regions represents localization with weak relation. The generated heatmap shows the portion with strong relation more clearly.

In Fig.~\ref{fig:heatmap_case8}, the detected B-Box was located at a little upper than the given B-Box. The red area of the heatmap was also at upper position. The detected B-Box is slightly larger than the given B-Box. The detected B-Boxes are almost same as the given B-Boxes except the different size. For better detection capability, the feature of the generated heatmap will be investigated with the medical specialists.

\section{Conclusion}
\label{sec:conclusion}
Deep learning is widely used in various kinds of research fields, especially image recognition. In our research, Adaptive DBN which can find the optimal network structure for given data was developed. The method shows higher classification accuracy than existing deep learning methods for several benchmark data sets. In this paper, Adaptive DBN was applied to not only classification task but also object detection task for finding B-Box. A new detection algorithm with the trained DBN network was proposed and probability of semantic object was visualized as heatmap. In the simulation results, the proposed method was evaluated on CXR8. The method showed higher performance for both classification and detection accuracy compared with some existing CNN methods. Our proposed method will be further improved for better detection capability by evaluating the method on the other large big data sets.

\section*{Acknowledgment}
This work was supported by JSPS KAKENHI Grant Number JP17J11178.

\end{document}